\documentclass[runningheads]{llncs}
\usepackage{graphicx}

\usepackage{tikz}
\usepackage{comment}
\usepackage{amsmath,amssymb} 
\usepackage{color}

\usepackage{placeins}
\usepackage{multirow}

\begin{document}
\pagestyle{headings}
\mainmatter
\def\ECCVSubNumber{100}  

\title{Cosine meets Softmax: A tough-to-beat baseline for visual grounding} 

\titlerunning{ECCV-20 submission ID \ECCVSubNumber} 
\authorrunning{ECCV-20 submission ID \ECCVSubNumber} 
\author{Anonymous ECCV submission}
\institute{Paper ID \ECCVSubNumber}

\titlerunning{CMSVG}
%
\author{Nivedita Rufus\and
Unni Krishnan R Nair\and
K. Madhava Krishna\and
Vineet Gandhi}
\authorrunning{N. Rufus et al.}
%
\institute{International Institute of Information Technology, Hyderabad, India \and
\email{{\{nivedita.rufus, unni.krishnan\}@research.iiit.ac.in}}\\
\email{{\{mkrishna, vgandhi\}@iiit.ac.in}}}
\maketitle

\begin{abstract}
In this paper, we present a simple baseline for visual grounding for autonomous driving which outperforms the state of the art methods, while retaining minimal design choices. Our framework minimizes the cross-entropy loss over the cosine distance between multiple image ROI features with a text embedding (representing the give sentence/phrase). We use pre-trained networks for obtaining the initial embeddings and learn a transformation layer on top of the text embedding. We perform experiments on the Talk2Car~\cite{Deruyttere_2019} dataset and achieve 68.7\% AP50 accuracy, improving upon the previous state of the art~\cite{deruyttere2020giving} by 8.6\%. Our investigation suggests reconsideration towards more approaches employing sophisticated attention mechanisms~\cite{hudson2018compositional} or multi-stage reasoning~\cite{deruyttere2020giving} or complex metric learning loss functions~\cite{musgrave2020metric} by showing promise in simpler alternatives.


\keywords{Autonomous Driving, Visual Grounding, Deep Learning}
\end{abstract}

\section{Introduction}
\label{sec:Introduction}
\begin{figure}[t]
    \centering
    \begin{tabular}[t]{  p{5.25cm} p{0.5cm} p{5.25cm} }
        \includegraphics[width=\linewidth]{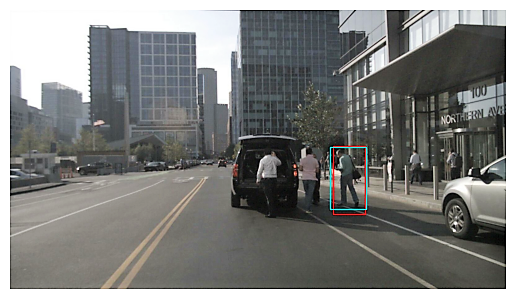}& 
          &
        \includegraphics[width=\linewidth]{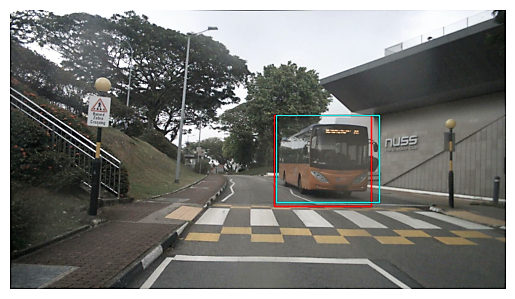}  \\ 
         {\scriptsize \textit{``My friend is the guy standing closest to the curb, next to that car in front of us. Pull over so he can get in".}} &  &
         {\scriptsize \textit{``Wait for this bus to pass, then turn left here. I mean right sorry".}} \\
    \end{tabular}
    \caption{ Illustrative examples of the visually grounded bounding boxes for the given commands. The ground truth bounding box is in red and the predicted bounding box is in cyan.
    }
    \label{fig:teaser}
\end{figure}

The launch of Level 5 autonomous vehicles requires the removal of all human controls. This warrants the need for a better system to direct the self-driving agent to perform maneuvers if the passenger wishes to do so. A convenient way to do so is through language based controls, where you can direct the car using natural language instructions. Sriram et al~\cite{8967929} were one of the first to address the problem, and they proposed a method to directly guide the car based on language instructions. However, the focus of the paper is on navigation aspect i.e. to employ 3D semantic maps to directly generate waypoints. They used minimal language instructions and their work cannot handle situations where the commands specify maneuvers which are described in rich detail and requires the maneuvers to be spatially constrained around them. We illustrate couple of such richly described examples in Figure \ref{fig:teaser}. 

To perform a requested maneuver the self-driving agent would be required to perform two steps. First, the agent should interpret the given command and ground it to some part of the visual space around it. Secondly, the agent has to devise a plan to perform the maneuver in this visual scene. Our paper focuses on this former step, or more concretely: given an image $I$ and a command $C$, the goal is to find the region $R$ in the image $I$ that $C$ is referring to. The task is commonly known as visual grounding or object referral and has been extensively studies in the past~\cite{plummer2015flickr30k,kazemzadeh2014referitgame,rohrbach2016grounding}. However, most of the previous studies uses generic datasets and our work focuses on the application of autonomous driving. More specifically, our work employs the recently proposed Talk2Car dataset~\cite{Deruyttere_2019} and our paper comes as an official entry to the Commands 4 Autonomous Vehicles (C4AV) competition 2020. The C4AV challenge reduces the complexity of the object referral task to the case where there is only one targeted object that is referred to in the natural language command. Here, the human occupant is expected to instruct the vehicle with natural language commands to perform actions subjected to predominantly the spatial information from the scene, such as \textit{`Park behind the white car'} or \textit{`Stop next to the pedestrian on the left'}.

There are two common methodologies to tackle the task of visual grounding. The first is to learn a coordinated representation for the image and text, such that matching pairs are highly correlated in the embedded space~\cite{wang2016learning,engilberge2018finding,karpathy2014deep}. Such methods are trained using margin based loss functions (contrastive, triplet etc.). While inference such models compute proposals and select the proposal with maximum correlation for the given text. The major challenge with these methods is difficulty in training due to the requirement of hard negative mining. The other line of approach employs attention modelling~\cite{rohrbach2016grounding,chen2017query}. In supervised setting, these methods learn a joint embedding space and employ a classification loss to ground the correct proposal~\cite{rohrbach2016grounding} or even regress the bounding box refinements around each proposal~\cite{chen2017query}. Some works employ multiple stages of attention based reasoning~\cite{caesar2019nuscenes}. In this work, we align these two streams of thoughts into a single design and employ a classification loss over cosine distance vector of multiple region proposal embeddings over a given sentence embedding. Albeit simple, our experiments show that such a minimal design can outperform state of the art approaches by a significant margin. Formally, our work makes following contributions: 

\begin{itemize}
    \item We propose a novel formulation which combines cross modal metric distances with a proposal level cross entropy loss on a given image. 
    \item We perform experiments on the Talk2Car~\cite{Deruyttere_2019} dataset and achieve 68.7\% AP50 accuracy on the test dataset provided by the Commands 4 Autonomous Vehicles (c4av) competition. Our work improves upon the previous state of the art~\cite{deruyttere2020giving} by 8.6\%
    \item We present extensive ablation study to motivate the choice of the base image and language embedding networks. The code for our work is publicly available at: https://github.com/niveditarufus/CMSVG
\end{itemize}



\section{Related Work}
\label{sec:Related Work}

Learning coordinated representations has been a common methodology in visual grounding literature. Given the image text feature, the aim here is to find projections of both views into a joint space of common dimensionality in which the correlation between the views is maximized. The correlation is often defined using cosine and euclidean metric. Early efforts rely on linear projections of the views/embeddings and use Canonical Correlation Analysis~\cite{plummer2015flickr30k}. More recent efforts employ deep networks on pre trained embeddings to learn a non linear projection~\cite{wang2016learning,engilberge2018finding,wang2018learning}. Most of these approaches employ the idea of region proposals and the goal is to find the best matching proposal in the image for a given phrase. The correlation networks is trained using a contrastive loss or a triplet loss. The requirement of hard negative mining remains a challenging factor in these approaches. A set of interesting proposal based efforts have been made in weakly supervised setting~\cite{karpathy2014deep,datta2019align2ground}, which use caption image alignment as the downstream task to guide the process of phrase localization. 

Attention modelling is another common methodology to solve the problem of visual phrase grounding. Rohrbach et al~\cite{rohrbach2016grounding} learn a joint embedding and give supervision over the attention to be given to each proposal. They also present an unsupervised formulation, using a proxy task of phrase reconstruction. They hypothesise that the phrase can only be reconstructed correctly if attention happens over correct proposal. Several approaches avoid object proposal and directly learn a spatial attention map instead. Javed et al.~\cite{javed2018learning} uses concept learning as a proxy task. Akbari et al.~\cite{akbari2019multi} address visual grounding by learning a multi-level common semantic space shared by the textual and visual modalities.

Our method also relates to the work by Hu et al.~\cite{hu2016natural}, which models this problem as a scoring function on candidate boxes for object retrieval, integrating spatial configurations and global scene-level contextual information into the network. Yu et al.~\cite{yu2018mattnet} address visual grounding using three modular components related to subject appearance, location, and relationship to other objects. Our work, moves away from instance level training to an image level training, where a proposal out of multiple candidates is picked using the cross entropy loss. 

Some efforts have been made specific to the autonomous driving scenario. The work by Sriram et al~\cite{8967929} directly regresses waypoints (the next point for the car to move) for the car based on natural language instructions. However, they work on a limited vocabulary of instructions.  The proposal of Talk2Car~\cite{Deruyttere_2019} dataset allows a richer exploration for the visual grounding task in autonomous driving scenarios. The work by Deruyttere et al. \cite{deruyttere2020giving} decomposes a query in a multistep reasoning process while continuously ranking 2D image regions during each step leading to low-scoring regions to be ignored during the reasoning process. This extends the work by Hudson et al.~\cite{hudson2018compositional} for multistep reasoning. Our work significantly reduces the complexity of the visual grounding architecture while outperforming several baselines and the state of the art model \cite{deruyttere2020giving}.

\begin{figure}[ht]
    \centering
    \begin{tabular}[t]{  p{5.25cm} p{0.5cm} p{5.25cm} }
        \includegraphics[width=\linewidth]{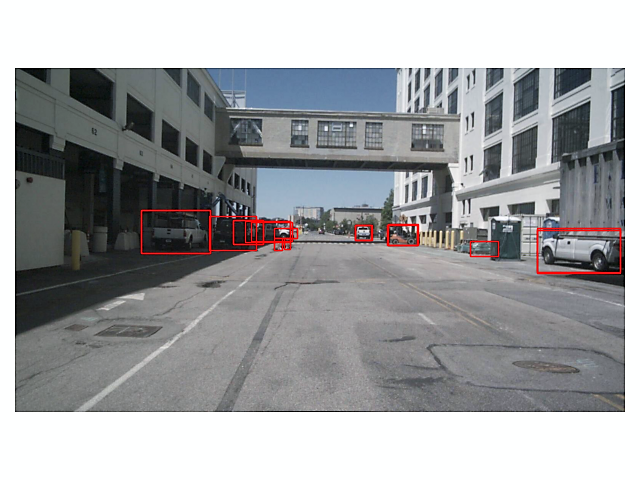}& 
          &
        \includegraphics[width=\linewidth]{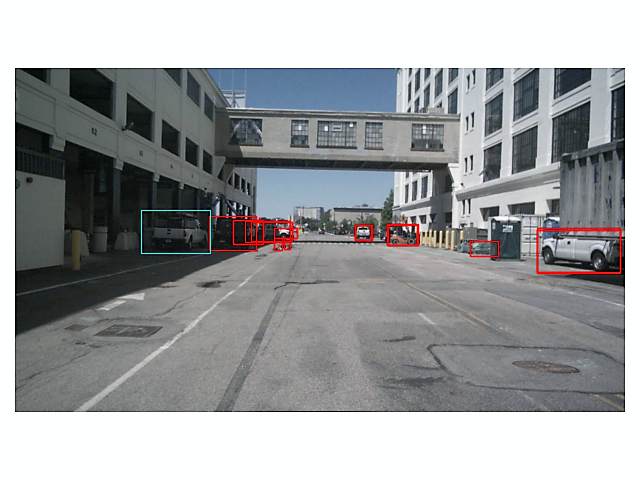}  \\ 
         {(a) All region proposals generated by the RPN} &  &
         {(b) The chosen object proposal(cyan) which is referred to by the command} \\
    \end{tabular}
    \caption{ Textual annotations are added to the nuScenes dataset \cite{caesar2019nuscenes} in the Talk2Car dataset. Each of the textual annotations refers to an object in the scene in the form of a command. The relevant object is selected from the proposals generated by the RPN. (a) shows the different bounding boxes generated by the RPN. The command corresponding to the (b) is \textit{{``Park next to the truck"}}}
    \label{fig:example}
\end{figure}

\section{The Proposed Model}
\label{sec:The Proposed Model}
\begin{figure}
    \centering
    \includegraphics[width=0.8\textwidth]{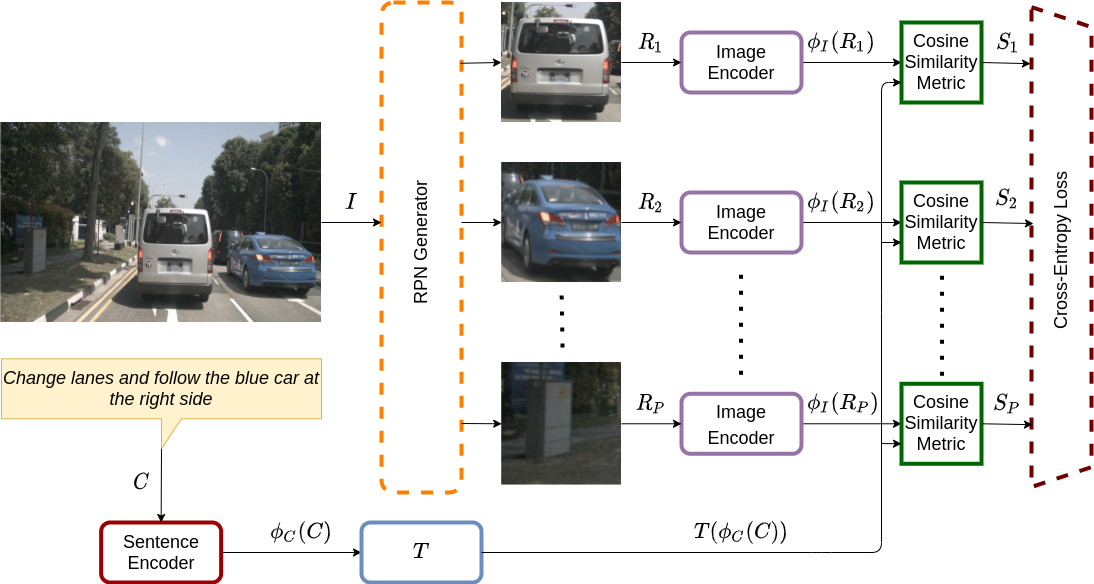}
    \caption{This figure shows an overview of the proposed method. We compute the cosine similarity between each of the encodings of the region proposals (from the image encoder) and the sentence embedding of the command (from the sentence encoder). The cosine similarity values are then considered as a score for how well the given bounding box fits the command. The criterion used for calculating the loss is a combination of Log-Softmax Loss and Negative Log Likelihood Loss.}
    \label{fig:overview}
\end{figure}
First, an object detector is used to extract region proposals from the input image. In this work, we use the pre-computed proposals provided to us by the competition organizers. These proposals are extracted using CenterNet \cite{duan2019CenterNet} as the RPN(\textbf{R}egion \textbf{P}roposal \textbf{N}etwork) and are illustrated in Figure \ref{fig:example}. Second, we match the region proposals with the command. In particular, we compute the cosine similarity values between the encodings of the region proposals and the transformed sentence embedding of the command. These similarity values are considered as a score for how well the bounding box fits the command.

Let us assume an object in the image $I$ is referenced by a command $C$. From the image $I$ we obtain a set of region proposals (each enclosing an object) $R_i$, ($i = 1, 2, ...P$),where $P$ is the number of proposals per image, generated by CenterNet \cite{duan2019CenterNet}. Each of the proposals is passed through an image encoder to obtain the feature vector $\phi_I(R_i)$. The transformation $T$ on sentence encoded feature vector of $C$ is given by $T(\phi_C(C))$, where $\phi_C(C)$ is the feature vector obtained from the sentence encoder. $S_i$ measures the feature similarity i.e. the cosine similarity between $\phi_I(R_i)$ and $T(\phi_C(C))$ for each proposal $R_i$, according to (\ref{eqn:cosine}).
\begin{equation}
\label{eqn:cosine}
    S_i = \frac{\phi_I(R_i)\cdot \
    T(\phi_C(C))}{|\phi_I(R_i)||T(\phi_C(C))|}, i = 1, 2, ...P
\end{equation}
To estimate the loss $L$, the criterion we employed that combines Log-Softmax loss and Negative Log Likelihood loss i.e. Cross-Entropy Loss, is given by (\ref{eqn:loss}).
\begin{equation}
    \label{eqn:loss}
    \begin{split}
        \alpha &= \frac{\exp{(S_g)}}{\sum_{i=1}^P\exp(S_i)} \\
         L &= -\log(\alpha)
    \end{split}
\end{equation}
where, $S_g$ is the score of the ground-truth proposal. We use pre-trained networks for obtaining the initial embeddings and learn a transformation layer on top of the sentence embedding. These model parameters are tuned using SGD with a small initial learning rate.

\section{Experiments and Results}
\label{sec:Experiments and Results}
\subsection{Dataset}
\label{subsec:dataset}
 The Talk2Car \cite{Deruyttere_2019} dataset contains 9,217 images from the nuScenes \cite{caesar2019nuscenes} dataset which have been taken in Sinagpore or Boston in different weather conditions as well as time conditions. They are annotated with natural language commands for self-driving cars, bounding boxes of scene objects, and the bounding box of the object that is referred to in a command. On average, a command and an image each contain respectively around 11 words and 11 objects from 23 categories. Train, validation contain respectively 8,349, 1,163 and 2,447 commands. In addition, the dataset consists of several smaller test sets, each of which evaluate specific challenging settings, like identifying far away objects. Another test set is made to assess how the model copes with cases which has multiple objects of the referred class in the visual scene. It also comprises of test case settings which evaluates how well a model is able able to deliver results with short and long sentences. An example of the Talk2Car dataset can be seen in Figure \ref{fig:teaser}.
 
 \subsection{Experimental details} 
\label{subsec:exp details}
We train the model parameters of the image encoder and the transformation block which is a fully-connected layer for 20 epochs using SGD with Nesterov momentum 0.9. The initial learning is set to 0.01 and decayed by a factor 10 every 4 epochs. We use batches of size 8 and a weight decay term 1e-4. The model can be trained in 1 hour on four Nvidia 1080ti GPUs. Our best model obtains 68.7\% AP50 on the test set.
 
 \subsection{Evaluation Metric} 
\label{subsec:eval}
We compare our method against some existing baseline approaches and also compare different different design choices for our model. This study was carried out on the Talk2Car \cite{Deruyttere_2019} dataset. The metric used for the comaparison is their respective AP50 scores on the dataset. AP50 score is defined as the percentage of the predicted bounding boxes that have an Intersection over Union (IoU) with the ground truth bounding boxes of over 0.5.

\subsection{Ablation Study} 
\label{subsec:ablation}
In this section we also show a detailed ablation study of how our model varies with various parameters and design choices. We studied the effect of the no. of proposals chosen has on the AP50 score in Table \ref{tab:proposals}. 
\begin{table}[!]
    \centering
    \begin{tabular}{|c|c|}
    \hline
         \textbf{No. of Proposals}& \textbf{AP50 on the Test data}  \\ \hline\hline
         8& 62.9\\ \hline
         16& 66.9\\ \hline
         32& 68.7\\ \hline
         48& 67.0\\ \hline
         64& 65.2\\ \hline
    \end{tabular}
    \caption{This table shows the variation of the AP50 scores on the test data with the number of region proposals generated by the RPN being used. The image encoder used is EfficientNet-B2 and the language encoder is STS RoBERTa(large).}
    \label{tab:proposals}
\end{table}
We also observe the trends of ResNet family of image encoders Table \ref{tab:resnet} and the EfficientNet family of image encoders in Table \ref{tab:eff-net}. We also study how the model performance varies with the different language encoders based on BERT \cite{devlin2018bert}, RoBERTa\cite{liu2019roberta} and DistilBERT\cite{sanh2019distilbert} trained on the STS\cite{reimers-2019-sentence-bert} benchmarking. 

\begin{table}[!]
    \centering
     \begin{tabular}{|l|p{1.5cm}|p{1.5cm}|p{1.5cm}|p{1.5cm}|p{1.5cm}|}
    \hline
     \multicolumn{1}{|l|}{\multirow{2}{*}{\textbf{Sentence Embedding}}} & \multicolumn{5}{c|}{\textbf{AP50 score on the Test data}}\\ \cline{2-6}
    & resnet-18& resnet-34& resnet-50& resnet-101& resnet-152\\ \hline\hline

    STS BERT(base)& 65.9& \textbf{66.6}& 66.2& 66.4& 64.9\\ \hline
    STS BERT(large)& 66.5& \textbf{67.6}& 66.7& 64.9& 65.1\\ \hline
    STS RoBERTa(base)& 66.5& 65.8& 64.4& 65.5& \textbf{66.8}\\ \hline
    STS RoBERTa(large)& 66.4& \textbf{67.6}& 67.0& 64.8& 65.6\\ \hline
    STS DistilBERT(base)& 66.1& \textbf{66.7}& 65.1& 66.6& 65.8\\ \hline
    \end{tabular}
    \caption{This table shows the variation of the AP50 scores on the the test data with various language encoders and the ResNet class of image encoders. Top 32 proposals generated by the RPN were used.}
    \label{tab:resnet}
\end{table}

Table \ref{tab:proposals} was generated from STS RoBERTa(large) as the language encoder and EfficientNet-B2 as the image encoder (our best performing model). From Table \ref{tab:proposals} we can infer that having too few proposals result in the object of interest being missed out. On the contrary, having too many proposals not only slows down the network but also decreases the performance. This can be attributed to the presence of multiple similar objects which contribute to the ambiguity in choosing the object that needs to be grounded. To get the best of both worlds, having 32 proposals seems to be a good trade-off.

\begin{table}[!]
    \centering
    \begin{tabular}{|l|p{1.5cm}|p{1.5cm}|p{1.5cm}|p{1.5cm}|p{1.5cm}|}
    \hline
     \multicolumn{1}{|l|}{\multirow{2}{*}{\textbf{Sentence Embedding}}} & \multicolumn{5}{c|}{\textbf{AP50 score on the Test data}}\\ \cline{2-6}
    & B0& B1& B2& B3& B4\\ \hline\hline

    STS BERT(base)& 65.8& \textbf{67.4}& 67.0& 66.3& 65.2\\ \hline
    STS BERT(large)& 67.3& \textbf{67.8}& 67.3& 66.4& 65.7\\ \hline
    STS RoBERTa(base)& 66.9& 66.5& \textbf{67.6}& 66.5& 65.8\\ \hline
    STS RoBERTa(large)& 68.2& 67.0& \textbf{68.7}& 66.1& 65.7\\ \hline
    STS DistilBERT(base)& \textbf{67.8}& 66.9& 67.1& 67.3& 65.8\\ \hline
    \end{tabular}
    \caption{This table shows the variation of the AP50 scores on the the test data with various language encoders and the EfficientNet class of image encoders. Top 32 proposals generated by the RPN were used.}
    \label{tab:eff-net}
\end{table}

All the results in Table \ref{tab:resnet} and Table \ref{tab:eff-net} were generated with 32 proposals. Though our best performing model turns out to be a combination of STS RoBERTa(large) as the language encoder and EfficientNet-B2 as the image encoder, STS DistilBERT(base) as the language encoder and EfficientNet-B0 as the image encoder yields quite close results at a much lower model complexity and higher inference speed. So, this would be the go-to option for real-time implementation. Another observation that can be made is that having deeper networks for the image encoders(both ResNet and EfficientNet) does not really improve the AP50 scores for the images. On the contrary, we get comparable or in some cases worse performance.

\subsection{Qualitative Results} 
\label{subsec: qualitative results}
\begin{figure}[t]
    \centering
    \begin{tabular}[t]{  p{5.25cm} p{0.5cm} p{5.25cm} }
        \includegraphics[width=\linewidth]{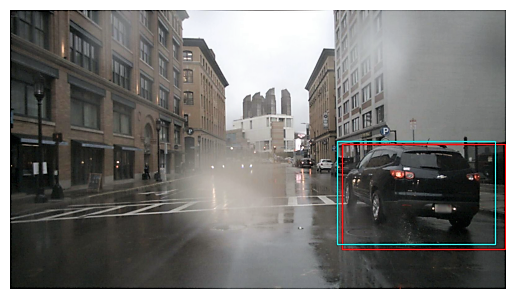}& 
          &
        \includegraphics[width=\linewidth]{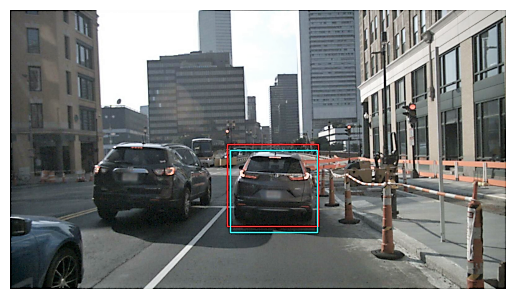}  \\ 
         {(a)\scriptsize \textit{``Follow the black car that is on your right".}} &  &
         {(b)\scriptsize \textit{``Do not move too close to this vehicle in front of us".}} \\
        \includegraphics[width=\linewidth]{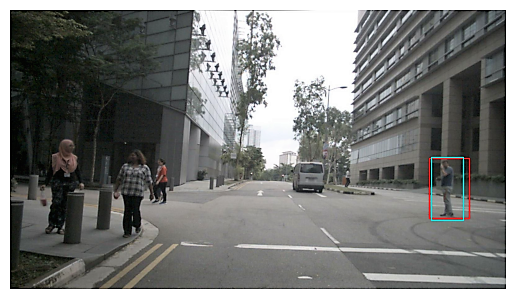}& 
          &
        \includegraphics[width=\linewidth]{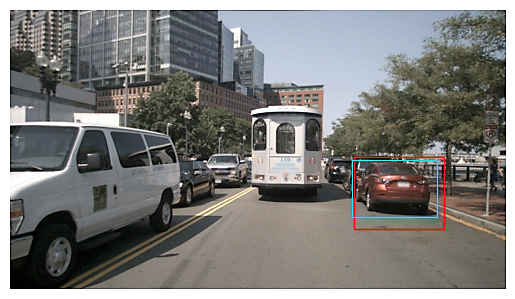}  \\ 
         {(c)\scriptsize \textit{``Slow down, because there is a man standing in traffic".}} &  &
         {(d)\scriptsize \textit{``Drop me off behind the red car".}} \\
    \end{tabular}
    \caption{ Illustrative examples of when our model predicts the grounded bounding boxes for the given commands correctly. The ground truth bounding box is in red and the predicted bounding box is in cyan.
    }
    \label{fig:corr}
\end{figure}

\begin{figure}[t]
    \centering
    \begin{tabular}[t]{  p{5.25cm} p{0.5cm} p{5.25cm} }
        \includegraphics[width=\linewidth]{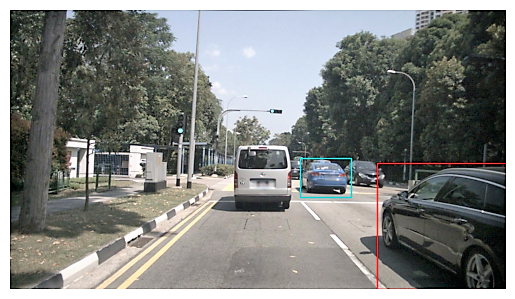}& 
          &
        \includegraphics[width=\linewidth]{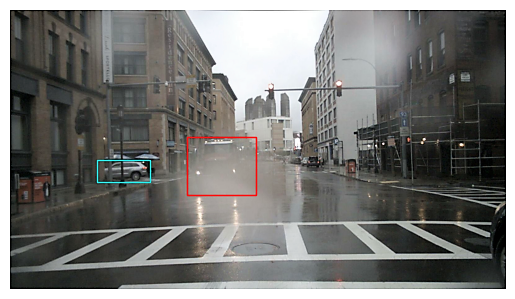}  \\ 
         {(a)\scriptsize \textit{``Accelerate and pass the car on the right, then change into the right lane".}} &  &
         {(b)\scriptsize \textit{``Let the red truck pass while continuing straight. Also use your windshield swipers man I do not see anything anymore".}} \\
        \includegraphics[width=\linewidth]{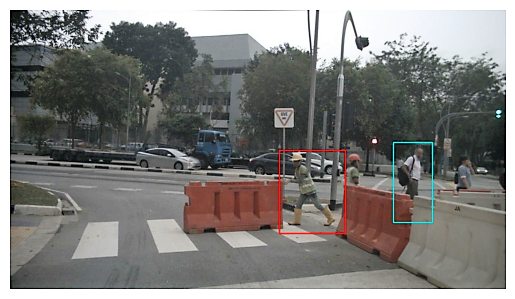}& 
          &
        \includegraphics[width=\linewidth]{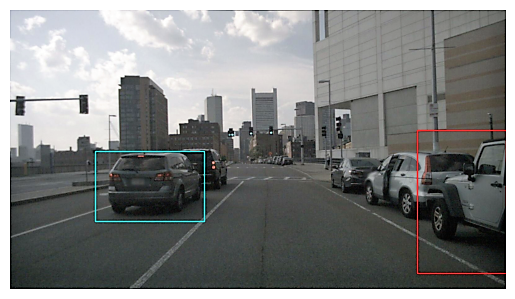}  \\ 
         {(c)\scriptsize \textit{``Do not proceed until this worker has cleared the intersection".}} &  &
         {(d)\scriptsize \textit{``Park next to the 4x4".}} \\
    \end{tabular}
    \caption{ Illustrative examples of when our model fails to predict the grounded bounding boxes for the given commands correctly. The ground truth bounding box is in red and the predicted bounding box is in cyan.
    }
    \label{fig:wrong}
\end{figure}

Some samples of how well our model performed can be visualized from Figure  \ref{fig:corr} and where it fails to deliver good results can be visualized in Figure  \ref{fig:wrong}. In the cases where it fails it can be seen that in Figure \ref{fig:wrong}b the truck is not visible because of occlusion due to fog, similarly in Figure \ref{fig:wrong}a the statement is ambiguous since both the cars are on the right and are candidates for an overtake. In Figure \ref{fig:wrong}c and Figure \ref{fig:wrong}d there are multiple objects from the same class, people and cars, respectively, so our model is not able to make such fine tuned distinction.

\subsection{Comparison With Existing Methods}
\label{subsec:comparision}
\begin{table}[!]
    \centering
    \begin{tabular}{|l|l|c|}
    \hline
         \textbf{S.No.}& \textbf{ Name of the Method }& \textbf{ AP50 on the validation data }  \\ \hline\hline
         1.& MAC& 50.51\\ \hline
         2.& STACK& 33.71\\ \hline
         3.& SCRC(top-32)& 43.80\\ \hline
         4.& A-ATT(top-16)& 45.12\\ \hline
         5.& MSRR(top-16)& 60.04\\ \hline
         6.& CMSVG(ours)& \textbf{68.20}\\ \hline
    \end{tabular}
    \caption{This table shows the comparison of the AP50 scores on the validation data with the existing state-of-the-art methods. The image encoder EfficientNet-B2 and the language encoder STS RoBERTa(large) is used for our model.}
    \label{tab:methods}
\end{table}
The improvements we got from the baseline for the competition \cite{v2020baseline} were majorly due to the embedding from the natural language encoding. Utilizing the embedding which used attention on the important parts of the sentence was a major contributor to the improvement in the performance. This might be because the average sentence length is 11 and there are sentences as big as 30+ words and attention helps the model to learn what is important in such a large command. We compare our model with the existing models in Table \ref{tab:methods}.

Our best performing model utilized EfficientNet-B2 \cite{tan2019efficientnet} as the image encoder for the top 32 proposals and RoBERTa \cite{liu2019roberta} pre-trained on STS benchmarking followed by a fully connected layer to calculate the cosine similarity score for the natural language encoding. We use this model to compare with the existing methods on the validation dataset of the Talk2Car \cite{Deruyttere_2019} dataset such as,
\begin{itemize}
    \item \textbf{MAC:} \cite{hudson2018compositional} uses a recurrent MAC cell to match the natural language command represented with a Bi-LSTM model with a global representation of the image. The MAC cell decomposes the textual input into a series of reasoning steps, where the MAC cell selectively attends to certain parts of the textual input to guide the model to look at certain parts of the image.
    
    \item \textbf{STACK:} The Stack Neural Module Network \cite{hu2018explainable} uses multiple modules that can solve a task by automatically inducing a sub-task decomposition, where each sub-task is addressed by a separate neural module. These modules can be chained together to decompose the natural language command into a reasoning process.
    
    \item \textbf{SCRC:} \cite{hu2016natural} A shortcoming of the provided baseline \cite{v2020baseline} is that the correct region has to be selected based on local information alone. Spatial Context Recurrent ConvNets match both local and global information with an encoding of the command.
    
    \item \textbf{A-ATT:} \cite{8578906} Formulates these challenges as three attention problems and propose an accumulated attention (A-ATT) mechanism to reason among them jointly. Their A-ATT mechanism can circularly accumulate the attention for useful information in image, query, and objects, while the noises are ignored gradually. 
    
    \item \textbf{MSRR:} \cite{deruyttere2020giving} Integrates the regions of a Region Proposal Network (RPN) into a new multi-step reasoning model which we have named a Multimodal Spatial Region Reasoner (MSRR). The introduced model uses the object regions from an RPN as initialization of a 2D spatial memory and then implements a multi-step reasoning process scoring each region according to the query.
\end{itemize}

\section{Conclusion and Future Work}
\label{sec:conclusion}
This paper formulates and evaluates a very simple baseline that beats the previous state of the art performance in this specific grounding task. 

We used pre-trained networks for obtaining the initial embeddings and learned a transformation layer on the text embedding. We performed experiments on the Talk2Car dataset and achieve 68.7\% AP50 accuracy, improving upon the previous state of the art by 8.6\%. Our investigation suggests reconsideration towards more simple approaches than employing sophisticated attention mechanisms or multi-stage reasoning, by showing promise in simpler alternatives.

We plan to integrate the depth data with LiDAR-camera calibration for further improvements. We also plan to work on the planning and navigation part and get it working in CARLA for a proof of concept work before integrating it with our in house autonomous driving vehicle SWAHANA. 

\section*{Acknowledgement}
This work was supported in part by Qualcomm Innovation Fellowship (QIF 2020) from Qualcomm Technologies, Inc. 

\bibliographystyle{splncs04}
\bibliography{egbib.bib}
\end{document}